\documentclass[conference]{IEEEtran}
\IEEEoverridecommandlockouts

\usepackage{amsmath,amssymb,amsfonts}
\usepackage{algorithmic}
\usepackage{graphicx}
\usepackage{textcomp}
\usepackage{xcolor}
\usepackage{booktabs}
\usepackage{multirow}
\usepackage{subcaption}
\usepackage{algorithm}
\usepackage{listings}
\usepackage{pifont}
\newcommand{\yes}{\textcolor{black}{\ding{51}}}
\newcommand{\no}{\textcolor{black}{\ding{55}}}

\definecolor{codeblue}{rgb}{0.25,0.5,0.5}
\definecolor{codesign}{RGB}{0, 0, 255}
\definecolor{codefunc}{rgb}{0.85, 0.18, 0.50}
\lstdefinelanguage{PythonFuncColor}{
  language=Python,
  keywordstyle=\color{blue}\bfseries,
  commentstyle=\color{codeblue},
  stringstyle=\color{orange},
  showstringspaces=false,
  basicstyle=\ttfamily\small,
  literate=
    {+}{{\color{codesign}+ }}{1}
    {-}{{\color{codesign}- }}{1}
    {*}{{\color{codesign}* }}{1}
    {compute_embed}{{\color{codefunc}compute\_embed}}{1}
    {latent_interpolation}{{\color{codefunc}latent\_interpolation}}{1}
    {compute_diversity}{{\color{codefunc}compute\_diversity}}{1}
    {compute_consistency}{{\color{codefunc}compute\_consistency}}{1}
    {auto_grad}{{\color{codefunc}auto\_grad}}{1}
    {norm_and_reg}{{\color{codefunc}norm\_and\_reg}}{1}
}
\usepackage[backend=bibtex,bibstyle=ieee,citestyle=numeric-comp]{biblatex}
\usepackage{hyperref}

\begin{document}

\title{Consistency-Preserving Diverse Video Generation}

\author{
    \IEEEauthorblockN{Xinshuang Liu, Runfa Blark Li, Truong Nguyen}
    \IEEEauthorblockA{\textit{University of California, San Diego, CA, USA}}
    \IEEEauthorblockA{\{xil235, rul002, tqn001\}@ucsd.edu}
}

\maketitle

\begin{abstract}
Text-to-video generation is expensive, so only a few samples are typically produced per prompt. In this low-sample regime, maximizing the value of each batch requires high \emph{cross-video diversity}.
Recent methods improve diversity for image generation, but for videos they often degrade \emph{within-video temporal consistency} and require costly backpropagation through a video decoder.
We propose a joint-sampling framework for flow-matching video generators that improves batch diversity while preserving temporal consistency.
Our approach applies diversity-driven updates and then removes only the components that would decrease a temporal-consistency objective.
To avoid image-space gradients, we compute both objectives with lightweight latent-space models, avoiding video decoding and decoder backpropagation.
Experiments on a state-of-the-art text-to-video flow-matching model show diversity close to strong joint-sampling baselines while substantially improving temporal consistency and color naturalness.
Our code is available at \url{https://github.com/XinshuangL/Diverse-Video}.
\end{abstract}


\section{Introduction}

\textbf{Background and goals.} Video generation is a fundamental problem in computer vision, enabling applications in media content creation~\cite{DBLP:conf/cvpr/HuangWCCYW25,DBLP:conf/iclr/YangTZ00XYHZFYZ25} and virtual reality~\cite{DBLP:conf/cvpr/YuDHF025,DBLP:conf/iclr/0003TXFDF0025}. However, generating videos is computationally expensive, limiting the number of samples that can be produced under a fixed computation budget. To maximize the utility of each generation, we aim to jointly sample a batch of \textbf{diverse} videos. Unlike images, videos also demand \textbf{temporal consistency}: frames within each video must remain coherent. Thus, our goal is twofold: (i) ensure high diversity across generated videos, and (ii) maintain temporal consistency within each video.

\textbf{Our approach.} We propose a flow-matching sampling framework that targets both goals. Inspired by diverse image generation methods~\cite{DBLP:conf/iclr/CorsoXBBJ24,DBLP:conf/cvpr/MorshedB25}, we encourage diversity by using the gradient of a batch diversity objective to push samples apart (a \emph{diversity gradient}) during the flow-matching sampling process~\cite{DBLP:conf/iclr/LipmanCBNL23}. For consistency, we compute a temporal-consistency objective and remove from the diversity gradient any component that would decrease this objective (using a \emph{consistency gradient}). This regulation protects temporal consistency while retaining diversity updates that are neutral or beneficial to consistency.

\textbf{Latent-space models.} Prior diverse image generation methods~\cite{DBLP:conf/iclr/CorsoXBBJ24,DBLP:conf/cvpr/MorshedB25} compute diversity gradients in image space and backpropagate through the decoder. For videos, the higher dimensionality makes such gradient computations memory-intensive and often infeasible to perform in parallel. We instead train lightweight latent models so both diversity and consistency objectives can be computed without decoder passes.

\textbf{Contributions.} We introduce (1) a consistency-preserving joint sampling method for flow-matching video generators via gradient regulation, and (2) latent-space embedding and interpolation models that enable lightweight computation in latent space. Experiments demonstrate that our approach substantially improves temporal consistency and color naturalness while achieving diversity close to strong joint-sampling baselines.

\begin{figure*}[ht]
    \centering
    \includegraphics[width=\linewidth]{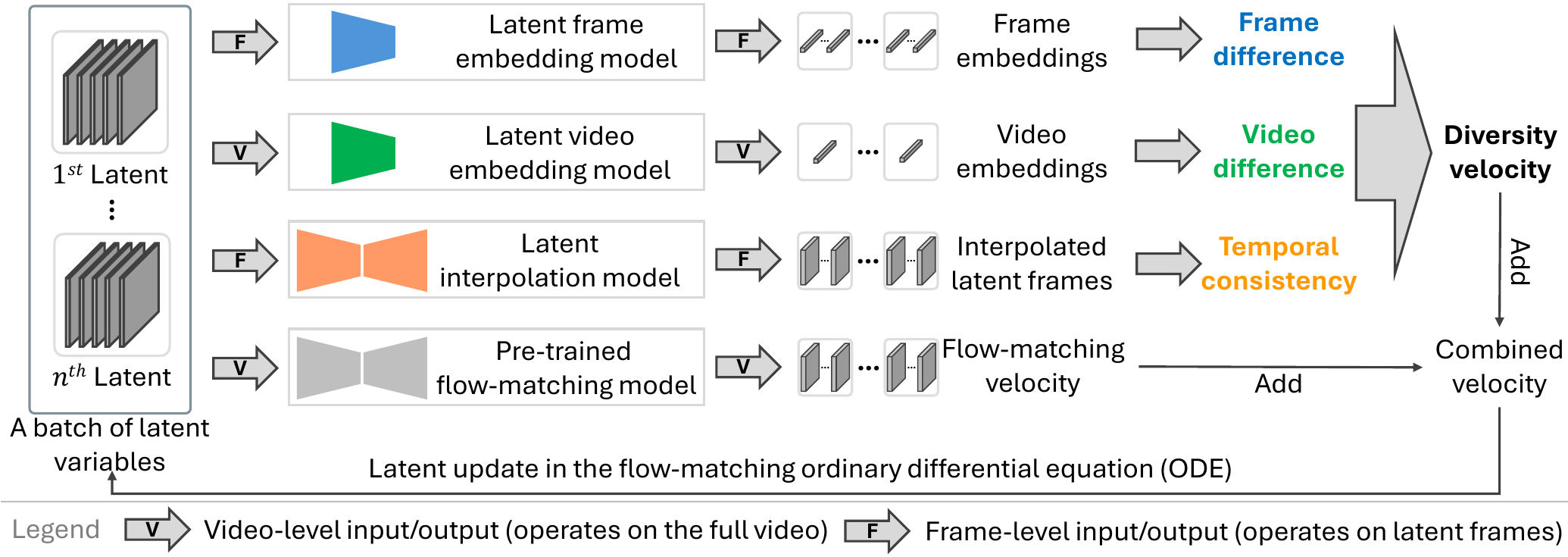}
    \caption{Joint video generation with enhanced cross-video diversity and preserved intra-video temporal consistency based on latent-space embedding and interpolation.}
    \label{fig:apply_models}
\end{figure*}

\section{Preliminaries and Related Work}
\label{sec:related}

\subsection{Image and Video Generative Models}

We briefly review generative models for visual data and motivate our choice of flow matching for video generation.
GANs~\cite{DBLP:conf/nips/GoodfellowPMXWOCB14} map noise to realistic images via adversarial training; VideoGAN~\cite{DBLP:conf/nips/VondrickPT16} extends GANs to videos, but adversarial training is prone to mode collapse.
Normalizing flows~\cite{DBLP:conf/iclr/DinhSB17} provide exact likelihoods via invertible mappings; VideoFlow~\cite{DBLP:conf/iclr/KumarBEFLDK20} applies this approach to video forecasting, but invertibility constraints can scale poorly to high resolution or long videos.
Diffusion models~\cite{DBLP:conf/nips/HoJA20} generate data by iterative denoising; video diffusion models~\cite{DBLP:conf/nips/HoSGC0F22} are effective but require many sequential steps per sample.

Flow matching~\cite{DBLP:conf/iclr/LipmanCBNL23} learns a time-dependent velocity field to transport a base distribution $p_0$ to the data distribution via an ODE. A sample $x_1$ is obtained by integrating
\begin{equation}
    \frac{\mathrm{d}}{\mathrm{d}t} x_t = v_\theta(x_t, t), \quad x_0 \sim p_0,
\end{equation}
from $t=0$ to $t=1$, where $v_\theta(x_t,t)$ is the learned velocity field.
Rectified flow~\cite{DBLP:conf/iclr/LiuG023} encourages nearly straight trajectories and has been adopted in text-to-video systems such as Wan~\cite{DBLP:journals/corr/abs-2503-20314}. We adopt flow matching for efficient, high-quality sampling under limited compute budgets.

\subsection{Joint Generation to Enhance Diversity}

Recent works improve diversity in diffusion and flow-matching \emph{image} generators by adding a diversity velocity to the sampling dynamics~\cite{DBLP:conf/iclr/CorsoXBBJ24,DBLP:conf/cvpr/MorshedB25}. These methods define a batch-level diversity objective $h(x_t^{(1:n)})$ over samples $x_t^{(1)},\dots,x_t^{(n)}$. For example,
\begin{equation}
D^{(i,i')} = \big\|F(x_t^{(i)}) - F(x_t^{(i')})\big\|^2_2, \quad
K^{(i,i')} = \frac{D^{(i,i')}}{\text{med}(D)}\,,
\end{equation}
where $\text{med}(D)$ is the median of all pairwise differences (treated as a constant to stabilize gradients). Following~\cite{DBLP:conf/cvpr/MorshedB25}, we evaluate $h$ on extrapolated end samples $\hat{x}_1^{(i)} = x_t^{(i)} + (1-t)\,v_\theta(x_t^{(i)},t)$.

To increase $h$ during sampling, these methods introduce a \emph{diversity velocity} $u$:
\begin{equation}
    u\big(x_t^{(i)}, x_t^{(-i)}, t\big) = \lambda_t\,\Phi\!\left(\nabla_{x_t^{(i)}}\, h(x_t^{(1:n)})\right),
\end{equation}
where $\lambda_t$ controls the strength and $\Phi$ optionally normalizes the gradient (we follow~\cite{liu2025importance} and normalize $u$ to match the norm of $v_\theta$). The $n$ samples evolve under coupled ODEs:
\begin{equation}
    \dot x_t^{(i)} = v_\theta\big(x_t^{(i)}, t\big) + u\big(x_t^{(i)}, x_t^{(-i)}, t\big), \quad
    x_0^{(i)} \overset{\text{i.i.d.}}{\sim} p_0,
    \label{eq:diversity_enhanced_ODEs}
\end{equation}
for $i=1,\dots,n$. While effective for images, these methods do not explicitly address temporal consistency or the cost of video-space gradients. Our work builds on this idea and tackles both issues.

\section{Methodology}
\label{sec:method}

\subsection{Consistency-Preserving Joint Video Generation}

Figure~\ref{fig:apply_models} provides an overview of our method. All guidance objectives are computed in latent space, avoiding decoder forward/backward passes.
We follow prior work to compute the diversity objective and add its gradient to the flow-matching dynamics via Eq.~(\ref{eq:diversity_enhanced_ODEs}). Specifically, we use a determinantal point process (DPP) objective~\cite{kulesza2012determinantal} (effective for image diversity~\cite{DBLP:conf/cvpr/MorshedB25}) over video-level and frame-level embeddings:
\begin{align}
e_v^{(i)} & = \text{Proj}\big(M_v(\hat{x}_1^{(i)}) \,\big|\, A_v b_v\big), \\ 
e_f^{(i,j)} & = \text{Proj}\big(M_f(\hat{x}_1^{(i,j)}) \,\big|\, A_f b_f \big), 
\end{align}
where $M_v$ and $M_f$ are latent-space video-level and frame-level embedding models, $A_v$ and $A_f$ are learned alignment matrices, and $\hat{x}_1 = x_t + (1-t)v_\theta(x_t,t)$ is the extrapolated terminal latent. Here, $i$ indexes the video sample and $j$ indexes the frame within video $i$; $b_v$ and $b_f$ are prompt embeddings from the reference video and frame encoders (VideoPrism-B~\cite{zhao2024videoprism} and CLIP~\cite{DBLP:conf/icml/RadfordKHRGASAM21}, respectively). The operation $\text{Proj}(a|b)$ removes the component of vector $a$ along $b$:
\begin{equation}
    \text{Proj}(a|b) = a - \frac{a \cdot b}{\|b\|_2^2}\, b \,.
\end{equation}
Using these embeddings, we compute video- and frame-level difference
matrices, normalize each by its median, and average:
\begin{align}
D^{(i,i')}_{v}
& = \big\|e_v^{(i)} - e_v^{(i')}\big\|_2^2,  \\
D^{(i,i')}_{f}
& = \frac{1}{T}\sum_{j=1}^{T} \big\|e_f^{(i,j)} - e_f^{(i',j)}\big\|_2^2, \\
K^{(i,i')} & = \frac{1}{2}\!\left(
  \frac{D^{(i,i')}_{v}}{\text{med}(D_v)} +
  \frac{D^{(i,i')}_{f}}{\text{med}(D_f)} \right)\,.
\end{align}
where $\text{med}(\cdot)$ is the median over off-diagonal entries, treated as a constant.
Using $K$, we compute the DPP diversity objective $O_d$ and its gradients with respect to each latent $x_t^{(i)}$:
\begin{equation}
    g_d^{(i)} = \nabla_{x_t^{(i)}} O_d\,.
\end{equation}
Similarly, we compute a consistency objective using a latent frame-interpolation model $M_c$:
\begin{equation}
    O_c = -\big\|\hat{x}_1 - M_c(\hat{x}_1)\big\|_2
\end{equation}
and its gradients:
\begin{equation}
    g_c^{(i)} = \nabla_{x_t^{(i)}} O_c\,.
\end{equation}
We regulate the diversity gradient by removing only the component that would decrease $O_c$. For each sample (dropping $(i)$ for clarity), let
\begin{equation}
    \alpha = \frac{g_d \cdot g_c}{\|g_c\|_2^2}, \quad g_\parallel = \alpha\, g_c, \quad g_\perp = g_d - g_\parallel \,.
\end{equation}
We then define the regulated diversity gradient as
\begin{equation}
    g_\mathrm{reg} = g_\perp + \max(\alpha, 0)\, g_c \,,
\end{equation}
which drops the negative projection of $g_d$ onto $g_c$. Since $g_c=\nabla O_c$, an update $g$ decreases $O_c$ to first order only if $g^\top g_c<0$. Thus, we keep $g_\perp$ and discard only the anti-aligned component of $g_d$ (when $\alpha<0$). We use $g_\mathrm{reg}$ in place of $g_d$ when forming the diversity velocity in Eq.~(\ref{eq:diversity_enhanced_ODEs}).
This is related to gradient regulation used to improve image quality~\cite{liu2025importance}, but here it preserves temporal consistency in video generation.

\subsection{Training Latent Embedding and Interpolation Models}

\begin{figure}[ht]
    \centering
    \includegraphics[width=\linewidth]{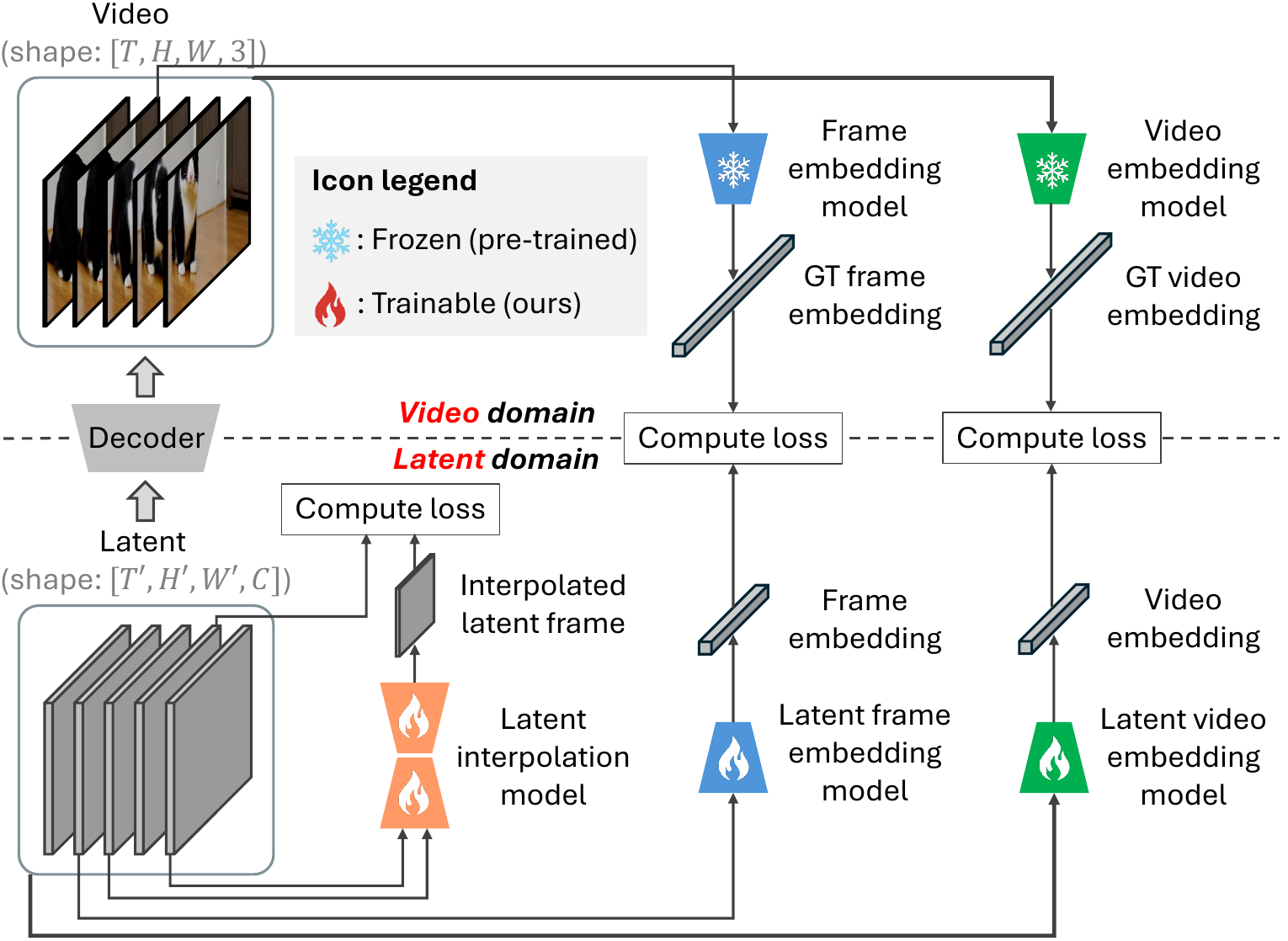}
    \caption{Illustration of training procedure for latent-space embedding and interpolation models.}
    \label{fig:train_models}
\end{figure}

We train small convolutional networks $M_v$, $M_f$, and $M_c$ as latent-space video embedding, frame embedding, and frame interpolation models. $M_v$ and $M_f$ mimic corresponding frozen encoders in video space (denoted $\tilde{M}_v$ and $\tilde{M}_f$), implemented with VideoPrism-B~\cite{zhao2024videoprism} and CLIP-B~\cite{DBLP:conf/icml/RadfordKHRGASAM21}.
As shown in Figure~\ref{fig:train_models}, during training, we decode a terminal latent $x_1$ to obtain a video $s$, compute reference embeddings $\tilde{e}_v$ and $\tilde{e}_f$ using frozen encoders, and train the latent models to match these embeddings when applied to the extrapolated latent $\hat{x}_1$ (using the same projection operation as in Section~\ref{sec:method}). Since each latent frame corresponds to four decoded video frames (after the first frame), we average the reference frame embeddings over each group of four frames.

We supervise $M_v$ so that pairwise dot products are preserved between latent space and video space:
\begin{equation}
L_s = \frac{1}{N^2} \sum_{i=1}^{N}\sum_{i'=1}^{N} \Big(e_v^{(i)} \cdot e_v^{(i')} - \tilde{e}_v^{(i)} \cdot \tilde{e}_v^{(i')}\Big)^2,
\end{equation}
where $N$ is the minibatch size. We further supervise the learned alignment $A_v$ by encouraging per-sample agreement:
\begin{equation}
L_p = \frac{1}{N} \sum_{i=1}^{N} \Big(1 - e_v^{(i)} \cdot (A_v \tilde{e}_v^{(i)})\Big).
\end{equation}
To regularize the embedding model output, we encourage correlation with a simple pooled latent statistic:
\begin{equation}
L_{\text{reg,m}} = 1 - \frac{1}{N} \sum_{i=1}^{N} e_v^{(i)} \cdot \text{mean}(\hat{x}_1^{(i)}),
\end{equation}
where $\text{mean}(\cdot)$ denotes spatial mean pooling over $\hat{x}_1^{(i)}$ to produce a vector. To regularize the alignment matrix, we add
\begin{equation}
    L_{\text{reg,p}} = \frac{1}{n_\text{latent} n_\text{image}}
        \sum_{k=1}^{n_\text{latent}} \sum_{l=1}^{n_\text{image}}
        \big| A_v^{(k, l)}\big|,
\end{equation}
where $n_\text{latent}$ and $n_\text{image}$ are the embedding dimensions in latent and decoded spaces. We also gradually mask out small-magnitude entries to encourage sparsity. The final loss for the video-level embedding model is
\begin{equation}
    L_\text{video} = \lambda_s L_s + L_p + L_{\text{reg,m}} + L_{\text{reg,p}},
\end{equation}
where $\lambda_s{=}10.0$. We use analogous losses for the frame embedding model and average the loss across temporal dimension.

For the latent frame interpolation model $M_c$, we train it to predict each latent frame $\hat{x}_1^{(i,j)}$ from its neighbors $\hat{x}_1^{(i,j-1)}$ and $\hat{x}_1^{(i,j+1)}$ using an MSE loss. To regularize, we additionally penalize deviation from the linear interpolation:
\begin{equation}
\big\| \hat{x}_{1,\text{pred}}^{(i,j)} - \tfrac{1}{2} (\hat{x}_1^{(i,j-1)} + \hat{x}_1^{(i,j+1)}) \big\|_2^2.
\end{equation}

\section{Experiments}
\label{sec:experiment}

\subsection{Experimental Setup}

\textbf{Video generation model and prompts.} We use Wan 2.1 t2v-1.3B~\cite{DBLP:journals/corr/abs-2503-20314} as the base text-to-video generator, generating videos using 50 flow-matching steps with diverse prompts:
\begin{itemize}
    \item $T_1$: ``A vehicle moves through an open landscape.''
    \item $T_2$: ``A pet plays on the floor.''
    \item $T_3$: ``An insect eats on a leaf.''
    \item $T_4$: ``An object emerges from the water.''
    \item $T_5$: ``A toy lights up in a dark room.''
    \item $T_6$: ``A tool operates on a workbench.''
    \item $T_7$: ``A fruit rolls across a table.''
    \item $T_8$: ``A plant sways in the wind outdoors.''
    \item $T_9$: ``A material reacts to its surroundings.''
    \item $T_{10}$: ``A group plays tennis on a court.''
\end{itemize}
We jointly generate 4 videos per prompt, repeating the generation 50 times to ensure evaluation accuracy.

\textbf{Training of the latent-space models.} We generate 100 training and 20 test videos for each prompt to train and evaluate the latent models. The embedding models are trained using samples in early, middle, and late stages (4{,}000 steps each). During the middle 4{,}000 steps, we gradually sparsify the projection matrices. The latent frame interpolation model is trained for 1{,}000 epochs.

\textbf{Evaluation metrics.} We measure cross-sample diversity using the Vendi score~\cite{DBLP:journals/tmlr/FriedmanD23} on video embeddings from VideoPrism-B~\cite{zhao2024videoprism} (\textbf{Vendi-v}) and frame embeddings from CLIP~\cite{DBLP:conf/icml/RadfordKHRGASAM21} (\textbf{Vendi-f}). For temporal consistency, we interpolate each frame from its neighbors using EDEN~\cite{DBLP:conf/cvpr/ZhangCZL00W25}, then compute the mean squared error (\textbf{MSE}) to the original frame (lower is better). We assess color naturalness using the Color Naturalness Index (\textbf{CNI})~\cite{DBLP:journals/cviu/HuangWW06}, averaged over frames, following~\cite{DBLP:journals/pr/YadavS24}.

\textbf{Baselines.} We compare against prior diversity-enhancement methods adapted to our setting. We implement their diversity objectives in latent space using the spatial mean of each latent (video-level) and latent frame (frame-level). We consider three diversity objective formulations: (i) a determinantal point process (\textbf{DPP}) objective~\cite{kulesza2012determinantal}, (ii) \textbf{DiverseFlow}~\cite{DBLP:conf/cvpr/MorshedB25} (which improves upon DPP with additional quality control), and (iii) \textbf{Particle Guidance}~\cite{DBLP:conf/iclr/CorsoXBBJ24} (which uses an exponential diversity objective). In all cases, the diversity gradient is applied via Eq.~(\ref{eq:diversity_enhanced_ODEs}) during flow matching.

\begin{table}[htbp]
\caption{Video generation results. Metrics are reported as mean(uncertainty) with 95\% confidence intervals; uncertainty aligns with the last digit of the mean.}
\begin{center}
\begin{tabular}{l|cccc}
\toprule
Method & Vendi-v $\uparrow$ & Vendi-f $\uparrow$ & MSE $\downarrow$ & CNI $\uparrow$ \\
\midrule
IID & 1.55(1) & 1.69(1) & 0.0010(1) & 0.67(1) \\
\hline
DPP & 1.61(1) & 1.81(1) & 0.0028(2) & 0.65(2) \\
Particle Guidance & 1.61(1) & 1.82(1) & 0.0028(2) & 0.65(2) \\
DiverseFlow & 1.61(1) & 1.81(2) & 0.0029(2) & 0.65(2) \\
Ours & 1.60(1) & 1.76(1) & 0.0019(1) & 0.68(1) \\
\bottomrule
\end{tabular}
\label{tab:main}
\end{center}
\end{table}

\begin{figure*}[ht]
    \centering

    \begin{subfigure}[t]{0.99\linewidth}
        \centering
        \includegraphics[width=\linewidth]{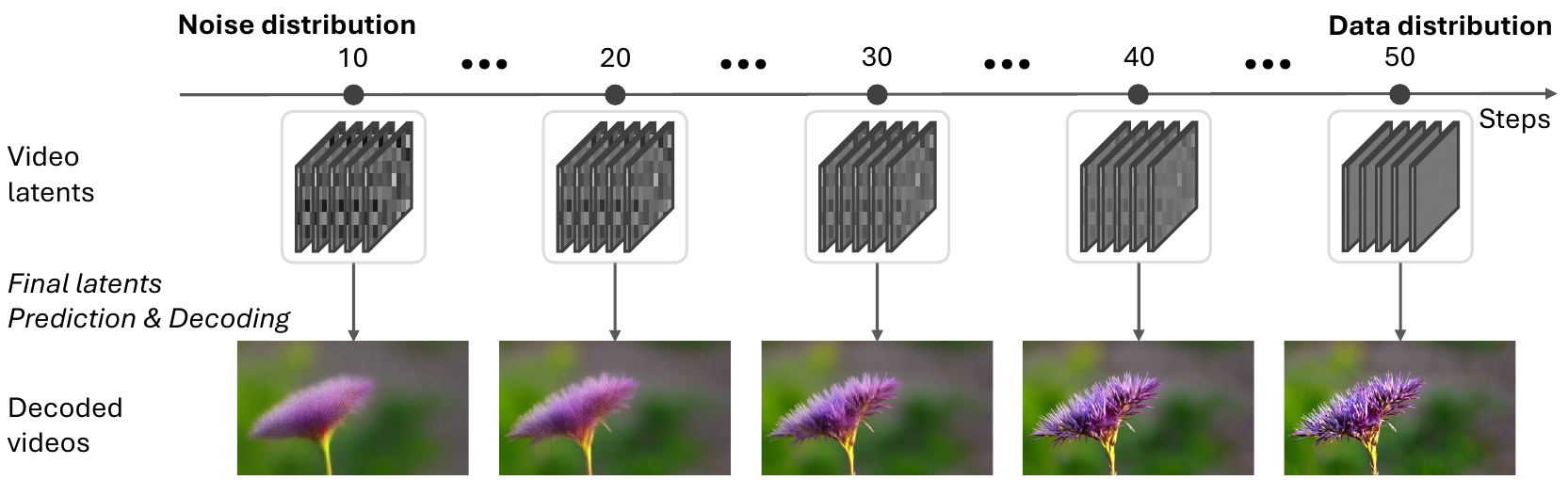}
        \caption{Extrapolation of $\hat{x}_1 = x_t + (1-t)v_\theta(x_t,t)$ from intermediate latent states and decoding them to video frames.}
    \end{subfigure}

    \par\medskip

    \newlength{\MyFigureRowSep}
    \setlength{\MyFigureRowSep}{0.8em}

    \begin{minipage}[t]{0.28\linewidth}
        \vspace{0pt}\centering

        \begin{subfigure}[t]{\linewidth}
            \centering
            \includegraphics[width=\linewidth]{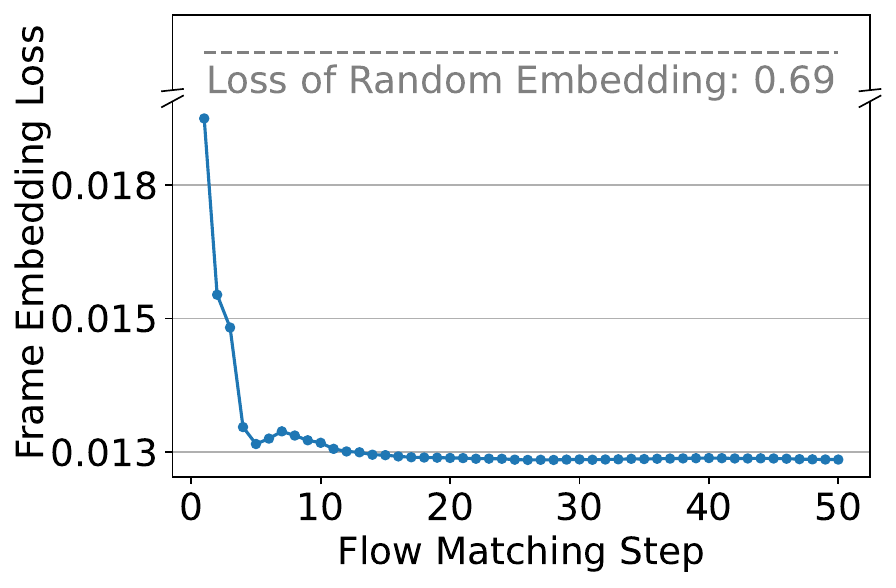}
            \caption{Frame embedding loss (ours).}
        \end{subfigure}

        \vspace{\MyFigureRowSep}

        \begin{subfigure}[t]{\linewidth}
            \centering
            \includegraphics[width=\linewidth]{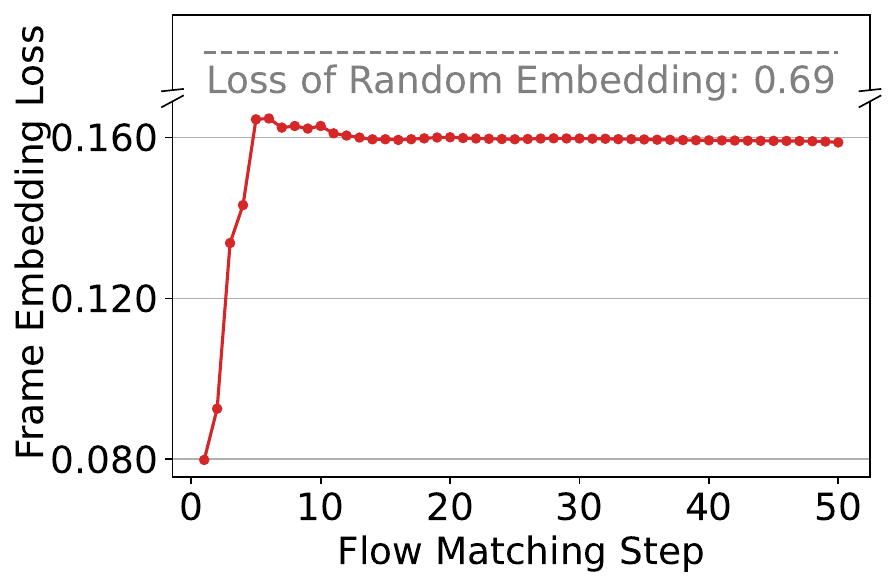}
            \caption{Frame embedding loss (baseline).}
        \end{subfigure}
    \end{minipage}
    \hfill
    \begin{minipage}[t]{0.28\linewidth}
        \vspace{0pt}\centering

        \begin{subfigure}[t]{\linewidth}
            \centering
            \includegraphics[width=\linewidth]{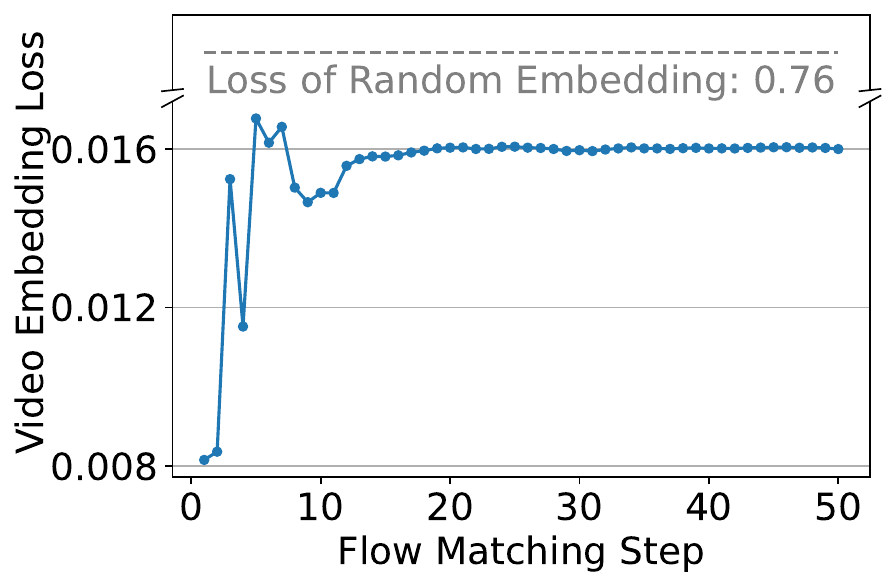}
            \caption{Video embedding loss (ours).}
        \end{subfigure}

        \vspace{\MyFigureRowSep}

        \begin{subfigure}[t]{\linewidth}
            \centering
            \includegraphics[width=\linewidth]{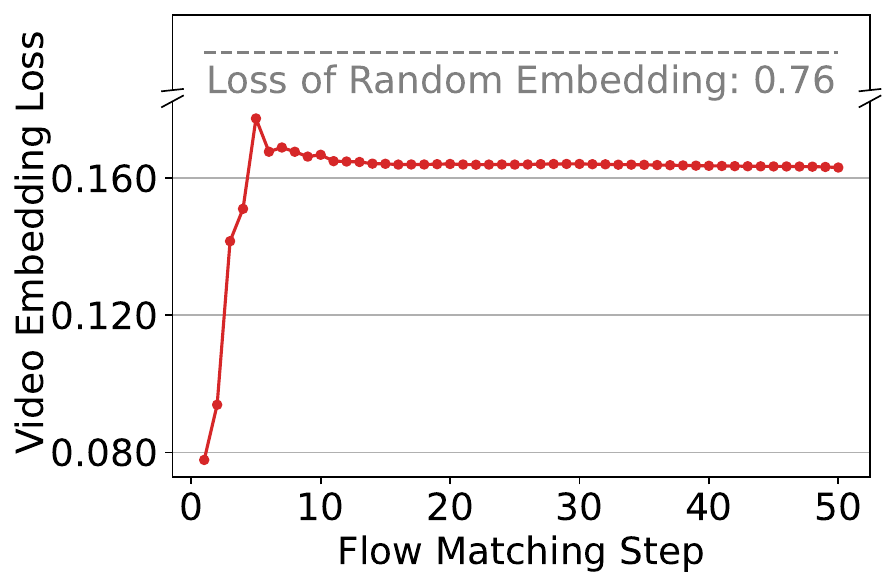}
            \caption{Video embedding loss (baseline).}
        \end{subfigure}
    \end{minipage}
    \hfill
    \begin{minipage}[t]{0.4\linewidth}
        \vspace{40pt}
        \centering
        \begin{subfigure}[t]{\linewidth}
            \centering
            \includegraphics[width=\linewidth]{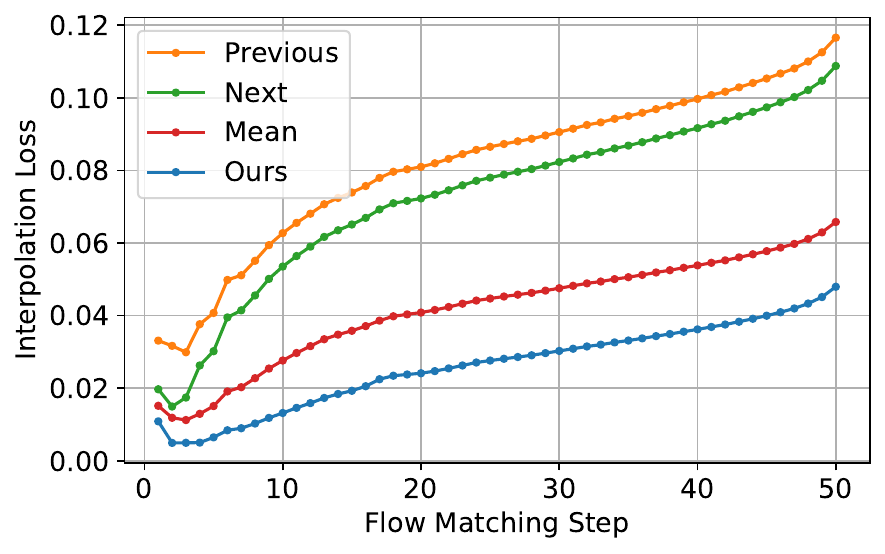}
            \caption{Frame interpolation loss across steps.}
        \end{subfigure}
    \end{minipage}

    \caption{Model behavior across flow-matching steps: 
    (a) Extrapolation of $\hat{x}_1 = x_t + (1-t)v_\theta(x_t,t)$ from intermediate latent states and decoding to video frames. 
    (b)--(e) Embedding alignment losses for our latent embedding models, compared to using a latent-mean baseline. 
    (f) Frame interpolation loss for $M_c$, compared to simple baselines (previous frame, next frame, and mean of both).}
    \label{fig:x1_prediction}
\end{figure*}

\subsection{Main Results}
Table~\ref{tab:main} presents results. All joint-sampling methods improve diversity over i.i.d.\ sampling, confirming the benefit of non-i.i.d.\ sampling. However, the baselines reduce temporal consistency (higher MSE) and color naturalness (lower CNI). Our method approximately keeps diversity while significantly improving temporal consistency and color naturalness.

\begin{table}[htbp]
\caption{Ablation results. Metrics are reported as mean(uncertainty) with 95\% confidence intervals.}
\begin{center}
\setlength{\tabcolsep}{4.5pt}
\begin{tabular}{cc|cccc}
\toprule
Diversity-v & ConsisReg & Vendi-v $\uparrow$ & Vendi-f $\uparrow$ & MSE $\downarrow$ & CNI $\uparrow$ \\
\midrule
\no & \no & 1.61(1) & 1.77(1) & 0.0017(1) & 0.68(1) \\
\no & \yes & 1.60(1) & 1.76(1) & 0.0016(1) & 0.68(1) \\
\yes & \no & 1.61(1) & 1.77(1) & 0.0021(1) & 0.68(1) \\
\yes & \yes & 1.60(1) & 1.76(1) & 0.0019(1) & 0.68(1) \\
\bottomrule
\end{tabular}
\label{tab:ablation}
\end{center}
\end{table}

\subsection{Ablation Study}
In Table~\ref{tab:ablation}, consistency-based regulation reduces MSE while approximately preserving diversity and color naturalness. Adding the video-level diversity term (Diversity-v) cannot improve the performance.

\subsection{Evaluating the Latent Space Models}
As shown in Figure~\ref{fig:x1_prediction}, flow matching gradually converts noise into a clean video. We evaluate model performance on predicted terminal latents extrapolated from different intermediate flow-matching steps. For embeddings, we use a simple latent-mean vector as a baseline and compute the MSE between similarity values in latent space and those computed in video space. Our latent models significantly outperform these baselines. For latent frame interpolation, we consider using the previous frame, the next frame, and the average of the two as baselines. The mean baseline performs best among these baselines, but our model outperforms it across all flow-matching steps. These results validate the effectiveness of our latent-space models and explain why they enable strong guidance without decoder backpropagation.

\section{Conclusion}
We presented a consistency-preserving joint sampling framework for flow-matching video generation that manages the diversity--consistency tradeoff. Our approach augments the flow-matching dynamics with a batch diversity objective, and applies a consistency-based regulation that removes only diversity updates that would decrease the consistency objective. To avoid the high computational cost of decoder forward and backward passes, we train lightweight latent-space models to compute these diversity and consistency objectives.
On Wan 2.1 t2v-1.3B, our method achieves cross-video diversity close to strong joint-sampling baselines while yielding better temporal consistency and higher color naturalness.

\printbibliography

\end{document}